%% file: main_paper.tex
\documentclass[nohyperref]{article}

\usepackage{microtype}
\usepackage{graphicx}
\usepackage{subfig}
\usepackage{adjustbox}
\usepackage{booktabs} 
\graphicspath{ {./images/} }
\usepackage{hyperref}
\input{math_commands.tex}

\usepackage{xcolor}

\usepackage[accepted]{icml2022}

\usepackage{amsmath}
\usepackage{amssymb}
\usepackage{mathtools}
\usepackage{amsthm}
\usepackage{cases}
\usepackage{graphicx}
\usepackage[capitalize,noabbrev]{cleveref}
\usepackage{hyperref}

\theoremstyle{plain}
\newtheorem{theorem}{Theorem}[section]

\theoremstyle{definition}
\newtheorem{definition}[theorem]{Definition}

\theoremstyle{remark}

\usepackage[textsize=tiny]{todonotes}

\icmltitlerunning{Improving Task-free Continual Learning by Distributionally Robust Memory Evolution}

\begin{document}

\twocolumn[
\icmltitle{Improving Task-free Continual Learning by\\
Distributionally Robust Memory Evolution}




\begin{icmlauthorlist}
\icmlauthor{Zhenyi Wang}{yyy}
\icmlauthor{Li Shen}{comp}
\icmlauthor{Le Fang}{yyy}
\icmlauthor{Qiuling Suo}{yyy}
\icmlauthor{Tiehang Duan}{meta}
\icmlauthor{Mingchen Gao}{yyy}
\end{icmlauthorlist}

\icmlaffiliation{yyy}{Department of Computer Science and Engineering, University at Buffalo, NY, USA}
\icmlaffiliation{comp}{JD Explore Academy, Beijing, China}
\icmlaffiliation{meta}{Meta, Seattle, WA, USA}

\icmlcorrespondingauthor{Zhenyi Wang}{zhenyiwa@buffalo.edu}
\icmlcorrespondingauthor{Li Shen }{mathshenli@gmail.com}
\icmlcorrespondingauthor{Mingchen Gao}{mgao8@buffalo.edu}

\icmlkeywords{Machine Learning, ICML}

\vskip 0.3in
]



\printAffiliationsAndNotice{} 

\input{abstract}

\input{introduction}
\input{relatedwork}

\input{method}
\input{experiment}

\input{conclusion}


\clearpage

\bibliography{example_paper}
\bibliographystyle{icml2022}

\newpage
\appendix
\onecolumn

\section{Baselines} \label{sec:baseline}
The detailed descriptions of baselines are the following:

\paragraph{Experience Replay (ER)} \cite{ERRing19}, which stores a small set of examples from previous tasks with reservoir sampling \cite{ERRing19}. At later time, we randomly sample a set of examples from the memory buffer to train with the received mini-batch data together to avoid forgetting.

\paragraph{Maximally Interfering Retrieval (MIR)} \cite{NIPS2019_9357}, the goal of MIR is to select the examples that are easily forgettable for replay. The idea of MIR is not
using randomly selected data from the memory buffer, but instead replaying the samples that
would be (maximally) interfered by the new received data. Following \cite{NIPS2019_9357}, we evaluate the model forgetting on 25 examples for Mini-ImageNet dataset, and 50 examples for other datasets.

\paragraph{AGEM} \cite{AGEM19}, AGEM tries to ensure that at every
training step the average episodic memory loss over the previous tasks does not increase. They project the gradient update direction such that they are less interfered with current data by projecting the gradient to the closest in L2 norm gradient that keeps the angle within 90 degree.

\paragraph{Gradient-Based Sample Selection (GSS-Greedy)} \cite{aljundi2019gradient} is to encourage diverse examples in the memory buffer. We use GSS-Greedy, which is efficient and performs the best. 

\paragraph{GMED} \cite{jin2021gradientbased}. GMED is the recent memory-replay methods, which edits the memory data so that they are harder to be memorized, shares the similar hypothesis as \cite{NIPS2019_9357}.

\section{ Lagrangian Duality Derivation}
\label{sec:duality}

The basic idea in Lagrangian duality  \cite{boyd2004convex} is to consider the constraints by augmenting the objective function with a weighted sum of the constraint
functions. We first convert the optimization Eq. (\ref{eq:taskfree}), Eq. (\ref{eq:KL}) and Eq. (\ref{eq:dot}) into the following optimization.

\begin{gather}
    \min_{\forall \vtheta \in \bm{\Theta}} \sup_{\mu\in \gP}  \mathbb{E}_{\mu} \gL(\vtheta, \vx, y) \\ \label{eq:KL2}
    \textrm{s.t. }  \gP = \{\mu: \gD(\mu||\pi) \leq \gD(\mu_0||\pi) \leq \epsilon \} \\ \label{eq:dot2}
    -\mathop{\mathbb{E}}_{ \vx \sim \mu, \vx^{\prime} \sim \mu_0} \nabla_{\vtheta}\gL(\vtheta, \vx, y) \cdot \nabla_{\vtheta}\gL(\vtheta, \vx^{\prime}, y) \leq -\sigma.
\end{gather}

Then, by Lagrangian duality \cite{boyd2004convex}, we can get the equivalent constrained optimization:

\begin{equation}
\begin{aligned} \label{eq:opt_apx}
    \min_{\forall \vtheta \in \bm{\Theta}} \sup_{\mu}  [\mathbb{E}_{\mu} \gL(\vtheta, \vx, y) - \gamma \gD(\mu||\pi) + \\
    \beta \mathop{\mathbb{E}}_{ \vx \sim \mu, \vx^{\prime} \sim \mu_0}  \nabla_{\vtheta}\gL(\vtheta, \vx, y) \cdot \nabla_{\vtheta} \gL(\vtheta, \vx^{\prime}, y)].
\end{aligned}
\end{equation}

where $\gamma$ and $\beta$ are the Lagrange multiplier associated
with the corresponding inequality constraints.

\section{More Experiments Results}
\label{sec:appresult}
\textbf{Hyperparameter Sensitivity Analysis}

We use ER+WGF-LD as ablation study, similar results for other memory evolution methods. We perform sensitivity analysis for the regularization weight $\beta$ of gradient dot product between perturbed memory data and raw data. The results are shown in Table \ref{tab:sensivitybeta}. In particular, we show the results that $\beta = 0.0$, i.e., without gradient dot product regularization. Also, we perform sensitivity analysis on memory evolution rate $\alpha$ in Table \ref{tab:alphaCIFAR10}, \ref{tab:alphaCIFAR100} and \ref{tab:alphaImageNet}.

\begin{table}[h]
\centering
\caption{ Sensitivity analysis of regularization weight $\beta$.}
\begin{adjustbox}{scale=0.85,tabular= lccc,center}
\begin{tabular}{lrrrrrrr|c|}
\toprule
\textbf{Hyperparameter}& & CIFAR-10 & CIFAR-100 & Mini-Imagenet\\
\midrule
$\beta = 0.0$ && $37.2\pm 1.7$  &  $21.2\pm 1.5$     &  $27.1\pm 1.4$  \\
\midrule
$\beta = 0.001$ && $37.3\pm 1.8$  &  $21.6\pm 1.2$    &  $27.5\pm 1.8$  \\
\midrule
$\beta = 0.003$ && $37.6\pm 1.5$ & $21.5\pm 1.3$  & $27.3 \pm 1.0$  \\
\midrule
$\beta = 0.005$ && $37.5\pm 1.1$  &$21.3\pm 1.1$    &  $27.6\pm 1.4$  \\
\bottomrule
\end{tabular}
\label{tab:sensivitybeta}
\end{adjustbox}
\end{table}

\begin{table}[H]
\centering
\caption{ Sensitivity analysis of evolution rate $\alpha$ on CIFAR10.}
\begin{adjustbox}{scale=0.85,tabular= lccc,center}
\begin{tabular}{lrrrrrrr|c|}
\toprule
& $\alpha = 0.005$ & $\alpha = 0.01$ & $\alpha = 0.03$\\
\midrule
 & $37.5\pm 1.2$  &  $37.6\pm 1.5$  &    $37.9\pm 1.6$    &    \\
\bottomrule
\end{tabular}
\label{tab:alphaCIFAR10}
\end{adjustbox}
\end{table}

\begin{table}[H]
\centering
\caption{ Sensitivity analysis of evolution rate $\alpha$ on CIFAR100.}
\begin{adjustbox}{scale=0.85,tabular= lccc,center}
\begin{tabular}{lrrrrrrr|c|}
\toprule
&  $\alpha = 0.01$ & $\alpha = 0.05$& $\alpha = 0.1$ &\\
\midrule
 &$21.6\pm 1.2$ & $21.5\pm 1.3$   &  $21.3\pm 1.2$     &    \\
\bottomrule
\end{tabular}
\label{tab:alphaCIFAR100}
\end{adjustbox}
\end{table}

\begin{table}[H]
\centering
\caption{ Sensitivity analysis of evolution rate $\alpha$ on miniImageNet.}
\begin{adjustbox}{scale=0.85,tabular= lccc,center}
\begin{tabular}{lrrrrrrr|c|}
\toprule
& $\alpha = 0.0001$  &$\alpha = 0.001$& $\alpha = 0.005$ &  \\
\midrule
 & $27.5 \pm 1.2$  & $27.3 \pm 1.0$  &  $26.9 \pm 1.5$    &    \\
\bottomrule
\end{tabular}
\label{tab:alphaImageNet}
\end{adjustbox}
\end{table}

\section{Foundations of Calculus of Variations} \label{sec:COV}

In calculus of variations, the first variation is defined as following:

\begin{definition}[First Variations] 
The first variation of a functional $F(\mu)$ is the functional at $\mu$
\begin{equation}
    \frac{\delta F}{\delta \mu}(\mu) = \lim_{\epsilon \to 0} \frac{F(\mu + \epsilon \psi)- F(\mu)}{\epsilon},
\end{equation}
where $\psi$ is arbitrary function.
\end{definition}

For more detailed background knowledge of calculus of variations, we recommend the reader refer to \cite{variation}.

\section{Derivations of the memory evolution methods} \label{sec:derivationsevolution}

Specifically, we derive the WGF-SVGD as an example:

\begin{gather*}
    \frac{\delta \gD(\mu||\pi)} {\delta \mu} =  \log \frac{\mu}{\pi} + 1; \\
    \frac{\delta V(u)}{\delta \mu}\!=\!U(\vx,\! \vtheta) \!=\! -\! \gL(\vtheta,\!\vx,\! y)\!-\!\beta \nabla_{\vtheta}\gL(\vtheta,\! \vx,\! y)\!\cdot\! \nabla_{\vtheta}\gL(\vtheta,\! \vx^\prime\!,\! y). 
\end{gather*}

We define the target worst-case evolved memory data distribution as $\pi \varpropto e^{-U}$ by the energy function $U(\vx,\! \vtheta)$ defined above.

We replace the Wasserstein gradient $\nabla_{W_2}F(\mu_t)$  
by the transformation $\gK_{\mu} \nabla_{W_2}F(\mu_t)$ under the integral operator $\gK_{\mu}f(\vx) = \int K(\vx, \vx^{\prime})f(\vx^{\prime})d\mu(\vx^{\prime})$;

The probability measure in the kernelized Wasserstein space actually follows the kernelized WGF \cite{liu2017stein}:
\begin{equation} \label{eq:kernelWGF12}
     \partial_t \mu_t = div(\mu_t \gK_{\mu_t} \nabla \frac{\delta F}{\delta \mu}(\mu_t)).
\end{equation}
Eq. (\ref{eq:kernelWGF12}) can be viewed as the WGF Eq. (\ref{eq:gf}) in RKHS. It indicates that the random variable $X_t$ which describes the evolved memory data at time $t$ evolves as the following differential equation \cite{liu2017stein, NEURIPS2020_SVGD}:

\begin{equation} \label{eq:WGFSVGD2}
   \frac{d X}{dt} = - [\gK_{\mu} \nabla \frac{\delta F}{\delta \mu}(\mu_t)](X).
\end{equation}

First, we derive the kernelized Wasserstein gradient of 

\begin{equation} \label{eq:kernelgradient}
\gK_{\mu_t} \nabla \log \frac{\mu_t}{\pi}(\vx):= \int K(\vx, \cdot) \nabla \log \frac{\mu_t}{\pi} =  \int K(\vx, \cdot) \nabla U d\mu_t - \int \nabla_2 K(\vx, \cdot)d\mu_t
\end{equation}

where,  we apply integration by parts in the second identity. $\nabla_2$ means the gradient of the kernel w.r.t. the second argument.

 It indicates that the random variable $X_t$ which describes the evolved memory data at time $t$ evolves as the following differential equation \cite{liu2017stein}:

We plug Eq. \ref{eq:kernelgradient} into Eq. \ref{eq:WGFSVGD2}, we can obtain the following equations.

\begin{equation} \label{eq:ODEevolve}
   \frac{d X}{dt} = -\int K(\vx, \cdot) \nabla U d\mu_t + \int \nabla_2 K(\vx, \cdot)d\mu_t 
\end{equation}

 If we discretize the above Eq. \ref{eq:ODEevolve} and view each datapoint in memory as one particle, we can obtain the following memory evolution update equation:
 
\begin{equation}
\small \label{eq:particlesteinGD}
    \vx_{t+1}^i -  \vx_{t}^i  = -\frac{\alpha}{N} \sum_{j=1}^{j= N}  [\underbrace{k(\vx_t^i, \vx_t^j) \nabla_{\vx_t^j} U(\vx_t^j, \vtheta)}_{\text{smoothed gradient}} +  \underbrace{\nabla_{\vx_t^j} k(\vx_t^i, \vx_t^j)}_{\text{repulsive term}}],
\end{equation}

\end{document}

%% file: math_commands.tex
\usepackage{amsmath,amsfonts,bm}

\def\eqref#1{equation~\ref{#1}}

\def\1{\bm{1}}

\def\vtheta{{\bm{\theta}}}

\def\vr{{\bm{r}}}

\def\vv{{\bm{v}}}

\def\vx{{\bm{x}}}

\def\vz{{\bm{z}}}

\def\mC{{\bm{C}}}
\def\mD{{\bm{D}}}

\def\mI{{\bm{I}}}

\def\mQ{{\bm{Q}}}

\DeclareMathAlphabet{\mathsfit}{\encodingdefault}{\sfdefault}{m}{sl}
\SetMathAlphabet{\mathsfit}{bold}{\encodingdefault}{\sfdefault}{bx}{n}

\def\gD{{\mathcal{D}}}

\def\gH{{\mathcal{H}}}

\def\gK{{\mathcal{K}}}
\def\gL{{\mathcal{L}}}
\def\gM{{\mathcal{M}}}

\def\gP{{\mathcal{P}}}

\def\sR{{\mathbb{R}}}

\def\sW{{\mathbb{W}}}



%% file: abstract.tex
\begin{abstract}
Task-free continual learning (CL) aims to learn a non-stationary data stream without explicit task definitions and not forget previous knowledge. The widely adopted memory replay approach could gradually become less effective for long data streams, as the model may memorize the stored examples and overfit the memory buffer. Second, existing methods overlook the high uncertainty in the memory data distribution since there is a big gap between the memory data distribution and the distribution of all the previous data examples. To address these problems, for the first time, we propose a principled memory evolution framework to dynamically evolve the memory data distribution by making the memory buffer gradually harder to be memorized with distributionally robust optimization (DRO). We then derive a family of methods to evolve the memory buffer data in the continuous probability measure space with Wasserstein gradient flow (WGF). The proposed DRO is w.r.t the worst-case evolved memory data distribution, thus guarantees the model performance and learns significantly more robust features than existing memory-replay-based methods.  Extensive experiments on existing benchmarks demonstrate the effectiveness of the proposed methods for alleviating forgetting. As a by-product of the proposed framework, our method is more robust to adversarial examples than existing task-free CL methods. Code is available on GitHub \url{https://github.com/joey-wang123/DRO-Task-free} 

\end{abstract}

%% file: introduction.tex
\section{Introduction}

Continual learning (CL) is to learn on a sequence of tasks without forgetting previous ones. 
Most CL methods assume knowing task identities and boundaries during training. 
Instead, this work focuses on a more general and challenging setup, i.e., task-free continual learning \cite{aljundi2019taskfree}. This learning scenario does not assume explicit task definition, and data distribution gradually evolves without clear task boundaries, making it applicable to a broader range of real-world problems.

\begin{figure}[h]
\centering
\subfloat[$t=t_0$ \\(easy to memorize and overfit)]{\includegraphics[width=0.15\textwidth]{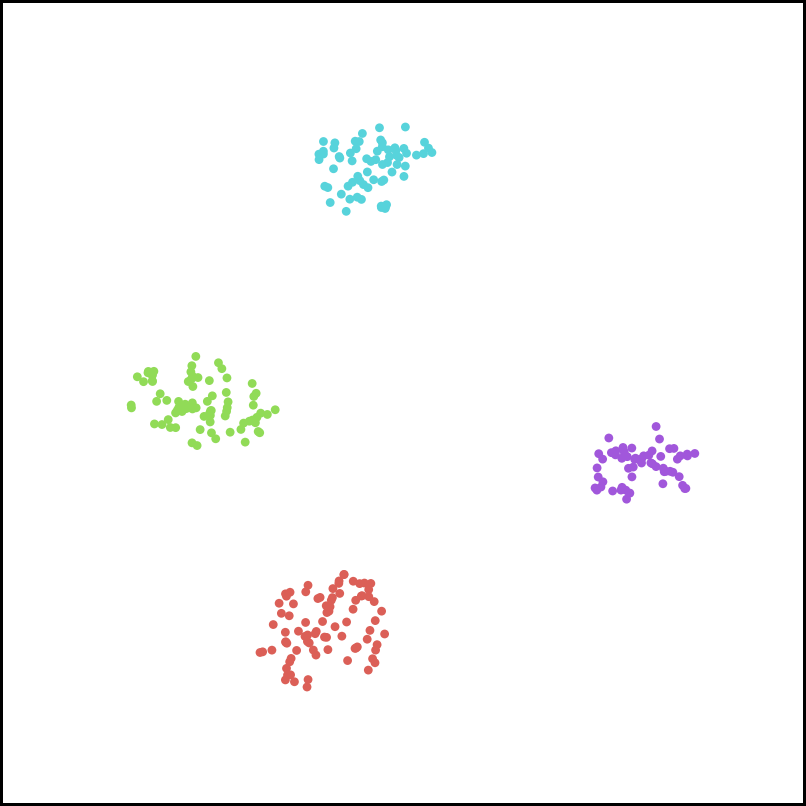}}  \,
\subfloat[$t=t_1$ \\(diverse and hard)]{\includegraphics[width=0.15\textwidth]{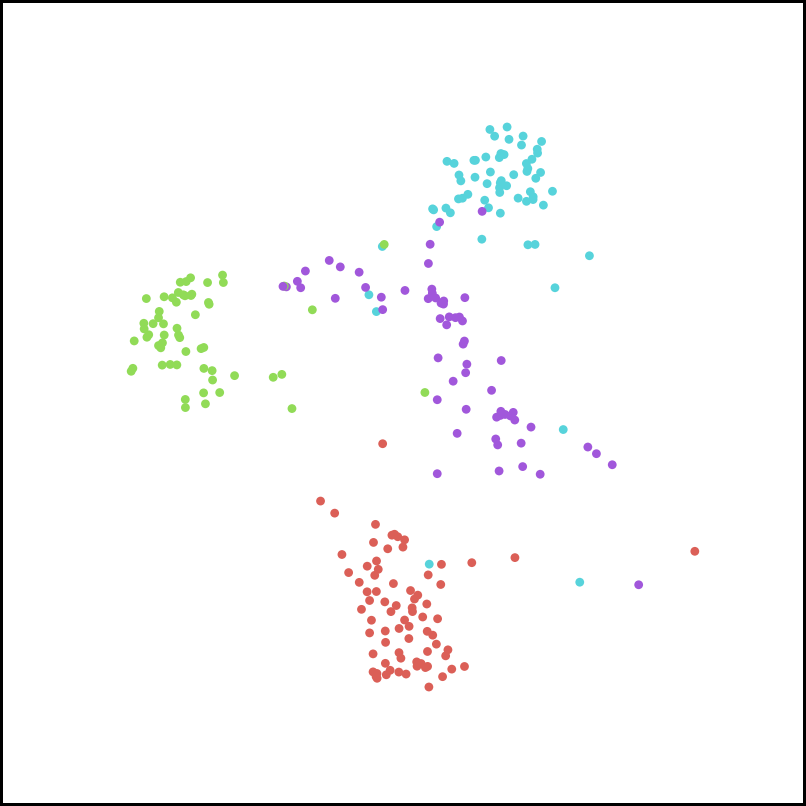}} \
\subfloat[$t=t_2$ \\(diverse and harder to classify)]{\includegraphics[width=0.15\textwidth]{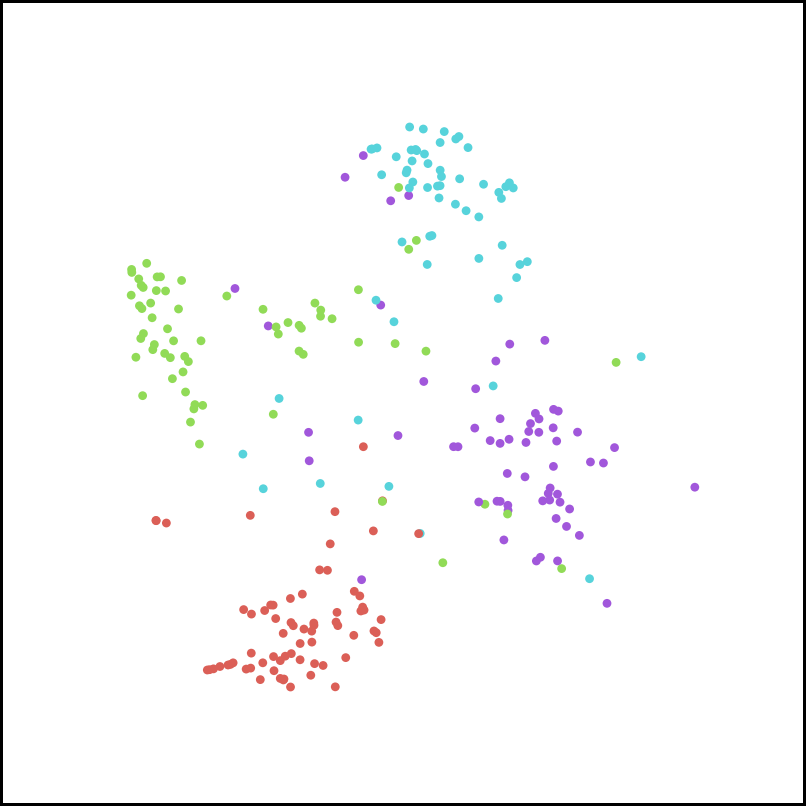}}
\caption{T-SNE visualization of evolved memory embedded by ResNet18 on CIFAR10 at different CL timestamps. Each dot represents a data point's feature extracted by the last layer of ResNet18. Each color denotes one class of memory data. Initially, the classes are very easy to memorize and overfit. Memory evolution makes the memory more diverse and harder to classify and memorize.} \label{fig:evolve1} 
\end{figure}

\begin{figure}[h]
\centering
\subfloat[Experience replay \\(raw memory data)]{\includegraphics[width=0.23\textwidth]{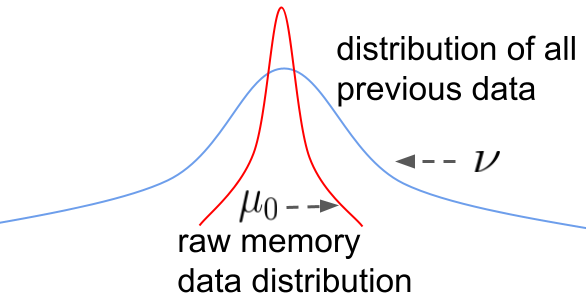}}  \,
\subfloat[Our method \\(evolved memory data)]{\includegraphics[width=0.23\textwidth]{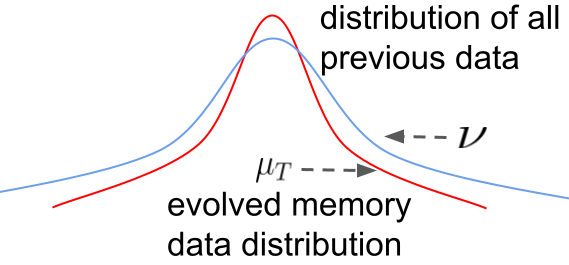}} \,
\caption{The blue line denotes the stationary distribution of all the previous data, and the red line denotes the memory buffer data distribution. (a): Standard experience replay (ER) (left) replays on the raw memory data. There is a big gap between the raw memory data distribution $\mu_0$ and the stationary distribution of all the data in the data stream $\nu$. (b): Our method (right) fills this gap by evolving the memory data distribution to narrow the gap between the evolved memory data distribution $\mu_T$ and $\nu$. }  \label{fig:evolve2}
\end{figure}

The current widely adopted memory replay approach stores a small portion of previous tasks in memory and replays
them with the new mini-batch data. 
The CL model would overfit the memory buffer, and this approach could be gradually less effective for mitigating forgetting as the model repeatedly learns the memory buffer \cite{jin2021gradientbased}, illustrated in Figure \ref{fig:evolve1} (a). The model could memorize the memory buffer, and previous knowledge will be quickly lost. Furthermore, there is a big gap between memory data distribution and the distribution of all the previous data examples. These methods ignore the high uncertainty of the memory data distribution since a limited memory buffer cannot accurately reflect the stationary distribution of all examples seen so far in the data stream, illustrated in Figure \ref{fig:evolve2} (a). Most existing CL works ignore this issue.

To address the above issues, we propose a distributionally robust optimization framework (DRO) to evolve the memory data distribution dynamically. This learning objective makes the memory data increasingly harder to be memorized and helps the model alleviate memory overfitting and improve generalization, illustrated in Figure \ref{fig:evolve1} (b) and (c).
In addition, the proposed DRO framework considers the underlying high uncertainty of memory data distribution. It evolves memory data distribution to fill the gap between the memory data distribution and the ideal stationary distribution of all the previous data, illustrated in Figure \ref{fig:evolve2} (b). We optimize the model performance under the worst-case evolved memory data distribution since we cannot know the exact distribution of all the previous examples. The model can thus learn substantially more robust features with performance guarantees than previous work. For the image domain, adversarial examples can have the same appearance as natural images but have different classification results. Our proposed DRO memory evolution framework is robust to such adversarial examples due to the worst-case performance optimization on the evolved memory data distribution. 

Formally, we first design energy functional that measures the degree of forgetting on a specific memory data distribution and the distributional distance between the evolved and raw memory distribution. The goal is to minimize the energy functional $F(\mu)$ for the worse-case performance probability measure (density) $\pi$ that is within the neighbor probability measures of the raw memory data probability measure $\mu_0$. We name this optimization as \textit{task-free DRO}. However, it is extremely challenging to solve this optimization problem in an infinite-dimensional probability measure space (function space) which contains an infinite number of probability measures. To efficiently solve the task-free DRO, we formulate it from a new continuous dynamics perspective and convert it into a gradient flow system. Specifically, the memory data distribution evolves in probability measure space as Wasserstein gradient flow (WGF), and model parameters evolve in Euclidean space. We name it as \textit{Dynamic DRO}, which makes it convenient to use \textit{function gradient-descent} in probability measure space to solve the task-free DRO. Also, it facilitates the derivation of different methods for memory data distribution evolution. 
We then introduce three specific memory evolution methods to solve the Dynamic DRO, including Langevin Dynamics \cite{SGLD11}, Stein Variational Gradient Descent \cite{liu2019stein}, and Hamiltonian flow \cite{ma2015complete, MCMCflow}. The proposed memory evolution framework is general, flexible, and easily extendable, with many potential extensions for future research.
We evaluate the effectiveness of the proposed framework by performing comprehensive experiments on several datasets and comparing them to various strong baselines.  
We summarize our contributions as follows:

\begin{itemize}
    \item We propose the first principled, general, and flexible \textbf{memory evolution} framework for task-free CL from the perspective of distributionally robust optimization, named \textit{task-free DRO}. Our proposed method is substantially more effective for mitigating forgetting.
    As a by-product of the proposed DRO framework, our method is more robust to adversarial examples than previous works. 
    
    \item We formulate the task-free DRO from a new continuous dynamics perspective and cast it as a gradient flow system, named \textit{Dynamic DRO}. We propose a family of memory evolution methods with different ways to efficiently solve the Dynamic DRO, opening up a new research direction for presenting new strategies to evolve the memory data.

    \item Extensive experiments on several datasets demonstrate the effectiveness of the proposed method for reducing forgetting and increasing the robustness to adversarial examples. Our framework is versatile and can be seamlessly integrated with existing memory-replay-based methods to improve their performance. 
\end{itemize}

%% file: relatedwork.tex
\section{Related Work}

\paragraph{Continual learning (CL)} aims to maintain previous knowledge when learning on sequential tasks with data distribution shift. 
Most existing CL methods \cite{lopezpaz2017gradient, nguyen2017variational, EWC16, zenke2017continual, oswald2019continual, Wang_2021_ICCV, saha2021gradient, pham2021contextual, Wang_2022_CVPR} are only applicable to the task-aware setting, where there are clear task definitions and boundaries among the sequentially learned task sequences.
Task-free continual learning \cite{he2019task, zeno2019task, aljundi2019taskfree, pmlr-v119-chrysakis20a} instead focuses on the more general case where the data distribution could change arbitrarily without explicit task splits. This learning scenario has become increasingly important due to the broader application scenarios, and more challenging problem nature \cite{aljundi2019taskfree, lee2020neural}.  

\paragraph{Task-free CL} Existing approaches for task-free CL can be categorized into two classes. The first (majority) one is 
memory-based methods \cite{aljundi2019gradient, NIPS2019_9357}, which store a small number of data from the previous data stream and replay them with the new mini-batch data later. The second type is the expansion-based method, such as CN-DPM \cite{lee2020neural}. However, CN-DPM needs to increase the memory and computation overhead with the network structure's expansion and the increase of the network parameters. 
Hence, this work focuses on memory-replay-based methods without expanding the network structure since it is simple and effective. Most existing works in this category \cite{ERRing19, AGEM19} directly perform replay on the raw data without any adaptation. MIR \cite{NIPS2019_9357} proposes to replay the samples with which are most interfered. GEN-MIR \cite{NIPS2019_9357} further uses generative models to synthesize the memory examples. The \textit{heuristic} method, GMED \cite{jin2021gradientbased}, proposes to edit memory examples so that the examples are more likely forgotten and harder to be memorized. Our methods share similar motivations with GMED, which \textit{individually} edits the memory data without considering memory data distribution uncertainty and population-level statistics. In contrast, our DRO framework focuses on \textit{population-level and distribution-level evolution}. Orthogonal to our work, Gradient-based Sample Selection (GSS) \cite{aljundi2019gradient} focuses on \textit{storing diverse examples}. These methods lack theoretical guarantees. In contrast, our framework is principled and focuses on evolving memory data distributions.

\paragraph{Distributionally robust optimization (DRO)}  is an effective optimization framework to handle decision-making under uncertainty \cite{rahimian2019distributionally}. The basic idea of DRO is first to construct a set of probability distributions as an ambiguity set and minimize the worst-case performance in this ambiguity set, thus guaranteeing the model performance. There are various applications of DRO in machine learning problems, including
tackling the group-shift \cite{sagawa2020distributionally}, subpopulation shift \cite{zhai2021doro}, and class imbalances \cite{imbalanceclass}.

To our best knowledge, our work is the first principled method with DRO for memory evolution in task-free CL, named task-free DRO. It dynamically evolves memory data distributions to avoid forgetting and learn robust features. In addition, we formulate the proposed task-free DRO from a new continuous dynamics perspective, making it convenient to handle the evolving memory data distribution with different evolution dynamics. Besides, we propose three ways to evolve the memory data, opening up a new research direction to explore more effective memory evolution methods.

%% file: method.tex
\section{Method}
In this section,  we first present the problem setup in Section \ref{sec:problem} for the task-free CL setting. Then, we propose our task-free DRO framework in Section \ref{sec:DRO}. Next, we view the task-free DRO from a new continuous dynamics perspective and formulate the task-DRO in an equivalent gradient flow system, named Dynamic DRO, in Section \ref{sec:dynamicdro}. In Section \ref{sec:WGF3}, we present three specific memory evolution methods to solve the Dynamic DRO efficiently.

\subsection{Problem Setup}
\label{sec:problem}
A sequence of mini-batch labeled data $(\vx_k, y_k, h_k)$ sequentially arrives at each timestamp  $k$ and forms a non-stationary data stream. Each data point is associated with a latent task identity $h_k$, where $\vx_k$ denotes the mini-batch data received at timestamp $k$, $y_k$ is the data label associated with $\vx_k$. According to \cite{aljundi2019taskfree}, a more general definition of task-free CL is that data distribution shift could happen at any time without explicit task splits. Our method can be directly applied to those more general cases as well. During both the training and testing time, the task identity $h_k$ is not available to the learner. At the same time, a small memory buffer $\gM$ is maintained, and replay the data in  $\gM$ when learning the new task to avoid forgetting the previously learned knowledge. The memory buffer $\gM$ is updated by reservoir sampling (RS), similar to \cite{riemer2018learning}. The goal is to learn a model $f(\vx, \vtheta)$ to perform well on all the tasks seen until timestamp $k$. 
Standard memory replay for CL \cite{ERRing19} is to optimize an objective under a known probability distribution $\mu_0$. Formally speaking, CL with standard memory replay can be expressed as:
\begin{equation}
\begin{aligned} \label{eq:replay}
    \min_{\forall \vtheta \in \bm{\Theta}} [\gL(\vtheta, \vx_k,  y_k) + \mathop{\mathbb{E}}_{\vx \sim \mu_0} \gL(\vtheta, \vx, y)], \\
\end{aligned}
\end{equation}
where $\vtheta$ are model parameters and $\vx$ is the raw memory buffer data with probability measure (density) $\mu_0$, i.e., $\forall \vx \in \gM, \vx \sim \mu_0$. $\gL(\vtheta, \vx,  y)$ is the loss function associated with the data $(\vx, y)$.  In the following, we temporally omit the term $\gL(\vtheta, \vx_k,  y_k)$ due to the fact that $(\vx_k,  y_k)$ is the mini-batch data arrived at timestamp $k$ and does not depend on the memory data distribution.

\subsection{DRO for Task-free CL} 
\label{sec:DRO}

Distributionally Robust Optimization (DRO) \cite{rahimian2019distributionally} is a systematic and elegant framework
to model the decision-making with ambiguity in the underlying probability distribution. For task-free CL, different from the standard memory-replay methods in Section \ref{sec:problem}, implicitly assuming $\mu_0$ is known. In contrast, our proposed DRO framework takes that the underlying actual probability distribution of memory data $\mu$ is \textit{unknown} and lies in an ambiguity set of probability
distributions. Modeling the memory data uncertainty is particularly useful when the memory buffer is small compared to the whole dataset since the memory has limited coverage to approximate the stationary distribution of all examples seen so far, illustrated in Figure \ref{fig:evolve2} (a). Thus, there is significant uncertainty in modeling the multi-task learning scenarios (the performance upper bound) with only a small memory buffer. The proposed DRO framework optimizes the worst-case performance in the ambiguity set of probability distributions since we cannot access the distribution of all the previous data. Therefore, it helps the model generalize to previous tasks since it can potentially narrow the gap between the memory data distribution and the distribution of all the previous data, illustrated in Figure \ref{fig:evolve2} (b). As a by-product of this optimization framework, it also helps learn features robust to data distribution perturbations. On the other hand, the memory buffer is updated slowly during CL (e.g., by reservoir sampling), and standard memory-replay repeatedly trains the memory buffer, and the CL model can easily overfit the memory buffer, as illustrated in Figure \ref{fig:evolve1} (a). Thus, the memory buffer could become less effective for mitigating forgetting. By optimizing the worst-case evolved memory data distribution at each iteration, our proposed DRO can also alleviate the memory overfitting problem by transforming the memory data to make them more difficult to memorize. This is illustrated in Figure \ref{fig:evolve1} (b) and (c).
Mathematically speaking, the proposed DRO for task-free CL can be expressed as:
\begin{gather} \label{eq:taskfree}
    \min_{\forall \vtheta \in \bm{\Theta}} \sup_{\mu\in \gP}  \mathbb{E}_{\mu} \gL(\vtheta, \vx, y) \\ \label{eq:KL}
    \textrm{s.t. }  \gP = \{\mu: \gD(\mu||\pi) \leq \gD(\mu_0||\pi) \leq \epsilon \}, \\ \label{eq:dot}
    \mathop{\mathbb{E}}_{ \vx \sim \mu, \vx^{\prime} \sim \mu_0} \nabla_{\vtheta}\gL(\vtheta, \vx, y) \cdot \nabla_{\vtheta}\gL(\vtheta, \vx^{\prime}, y) \geq \lambda,
\end{gather}
where the inner $\sup$ optimization is to gradually make the memory data distribution increasingly harder to be memorized. $\gP$ in Eq. (\ref{eq:KL}) denotes the ambiguity set of probability measures (distributions or densities) for the memory data distribution to characterize its uncertainty. One common choice to define $\gP$ is through Kullback-Leibler (KL) divergence. $\gD(\mu_0||\pi)$ denotes the KL divergence between probability measure $\mu_0$ and $\pi$, where $\pi$ is the target worst-case evolved memory data distribution, i.e., the probability distribution $\pi$ that achieves $\sup_{\mu\in \gP}  \mathbb{E}_{\mu} \gL(\vtheta, \vx, y)$. $\epsilon$ is a constant threshold to characterize the closeness between $\mu_0$ and $\pi$ to ensure the worst-case evolved memory data distribution $\pi$ does not deviate from the raw memory data distribution $\mu_0$ too much. Eq. (\ref{eq:dot}) constrains the gradient dot product between the worst-case evolved memory data distribution and raw memory data distribution, i.e., $\nabla_{\vtheta}\gL(\vtheta, \vx, y) \cdot \nabla_{\vtheta}\gL(\vtheta, \vx^{\prime}, y)$, to avoid the evolved memory data deviate from the raw memory data too much. Intuitively, if the gradient dot product is positive, it means the evolved memory data has a similar update direction compared to the raw data. $\lambda$ is a threshold to determine the constraint magnitude.

To solve the optimization, i.e., Eq. (\ref{eq:taskfree}-\ref{eq:dot}), the worst-case optimization that involves the $\sup$ optimization within the KL-divergence ball is generally computationally intractable since it involves the optimization over infinitely many probability distributions. To address this problem, by Lagrange duality \cite{boyd2004convex}, we convert Eq. (\ref{eq:taskfree}-\ref{eq:dot}) into the following unconstrained optimization problem (detailed derivations are put in Appendix \ref{sec:duality}):
\begin{equation}
\begin{aligned} \label{eq:opt}
    \min_{\forall \vtheta \in \bm{\Theta}} \sup_{\mu}  [\mathbb{E}_{\mu} \gL(\vtheta, \vx, y) - \gamma \gD(\mu||\pi) + \\
    \beta \mathop{\mathbb{E}}_{ \vx \sim \mu, \vx^{\prime} \sim \mu_0}  \nabla_{\vtheta}\gL(\vtheta, \vx, y) \cdot \nabla_{\vtheta} \gL(\vtheta, \vx^{\prime}, y)],
\end{aligned}
\end{equation}
Where $\gamma$ and $\beta$ control the magnitude of regularization. Since the KL-divergence term $\gD(\mu||\pi)$ is implicitly handled by gradient flow (discussed in the following sections), we set $\gamma =1$ for simplicity throughout this paper. The gradient of the third term (gradient dot product term) can be efficiently approximated by finite difference in practice. The optimization Eq. (\ref{eq:opt}) is still computationally hard to solve because the inner $\sup$ optimization is over probability measure space, which is an infinite-dimensional \textit{function space}. We name Eq. (\ref{eq:opt}) as \textbf{task-free DRO}.

\subsection{Task-free DRO: A Continuous Dynamics View} \label{sec:dynamicdro}

To make it tractable to solve the task-free DRO Eq. (\ref{eq:opt}),  we formulate it from a new continuous dynamics perspective. This new perspective brings significant advantages over directly solving Eq. (\ref{eq:opt}): 1) the optimization of Eq. (\ref{eq:opt}) over probability measure space can be converted into a continuous probability distribution evolution, equivalent as a WGF, enabling gradient-based solutions in \textit{probability measure space (function space)}; 2)
we can efficiently solve the WGF by various methods, which provide potential derivations of many different memory evolution methods.  
We then propose a novel solution by decomposing the task-free DRO Eq. (\ref{eq:opt}) into a gradient flow system. Specifically, we solve the inner $\sup$ optimization problem for memory evolution with WGF in Wasserstein space of probability measures and solve the outer optimization problem for the model parameters with gradient flow in Euclidean space. We convert the task-free DRO into a gradient flow system that alternately updates the memory evolution and model parameters.

Given a memory buffer at timestamp $k$, the raw memory data is $\gM = \{(\vx^1_0, y^1), (\vx^2_0, y^2), \cdots, (\vx^N_0, y^N)\}$, where $N$ is the memory buffer size. We perform a similar memory evolution procedure at each CL timestamp and omit the CL timestamp $k$ for notation clarity. We denote $\vx_t^i$ as the $i^{th}$  datapoint in the evolved memory buffer after $t$ evolution steps.
 The raw memory data is assumed to be i.i.d. sampled from the random variable $X_0$, i.e., $\{\vx_0^1, \vx_0^2, \cdots, \vx_0^N \} \sim X_0$. $X_0$ follows the probability distribution $\mu_0$, i.e.,  $X_0 ~ \sim \mu_0$. At evolution time $t$, the memory data $\gM$ is evolved as random variable $X_t$, i.e., $\{\vx_t^1, \vx_t^2, \cdots, \vx_t^N \} \sim X_t$ and probability distribution of random variable $X_t$ follows the probability measure $\mu_t$, i.e., $X_t ~ \sim \mu_t$. The empirical measure on the evolved memory buffer at time $t$ is defined as $\hat{\mu}_t = \frac{1}{N} \sum_{i=1}^{i=N}\delta (\vx^{i}_t)$ and $\delta$ is the Dirac measure. We model the memory evolution process as continuous WGF in probability measure space, i.e., use $(\mu_t)_{t\geq 0}$ to model the probability distribution evolution of memory data. The evolving $(\mu_t)_{t\geq 0}$ will in turn determine the memory evolution process $(X_t)_{t\geq 0}$ in Euclidean space.

Below, we define and present the gradient flow in Wasserstein space.
Let $\gP_2 (\sR^d)$ denote the space of probability measures on $\sR^d$ with finite second-order moments. Each element $\mu \in  \gP_2 (\sR^d)$ is a probability measure represented by its density function $\mu:\sR^d \to \sR$ with respect to Lebesgue measure $dx$. The Wasserstein distance between two probability measures $\mu_1, \mu_2 \in \gP_2 (\sR^d)$ is defined as:
\begin{equation*}
    W_2(\mu_1, \mu_2) = \left( \min_{\omega \in \prod(\mu_1, \mu_2)} \int ||\vx - \vx^{\prime}||^2 d\omega(\vx, \vx^{\prime})) \right)^{1/2},
\end{equation*}
where $\prod(\mu_1, \mu_2) = \{\omega |\omega(A \times \sR^{d}) = \mu_1(A), \omega(\sR^{d}\times B) = \mu_2(B)\}$. 
  $\omega$ is the joint probability measure with marginal measure of $\mu_1$ and $\mu_2$ respectively. Thus, $\sW^2 = (\gP_2 (\sR^d), W_2)$ forms a metric space.

\begin{definition}[Wasserstein Gradient Flow] \cite{gradientflow}.
Suppose we have a Wasserstein space $\sW^2 = (\gP_2 (\sR^d), W_2)$. A curve of $(\mu_t)_{t\geq 0}$ of probability measures is a Wasserstein gradient flow for functional $F$ if it satisfies
\begin{equation}\label{eq:LD}
   \partial_t \mu_t = \nabla_{W_2}F(\mu_t) := div\left(\mu_t \nabla \frac{\delta F}{\delta \mu}(\mu_t)\right)
\end{equation}
\end{definition}
where $: =$ means defined as, and $div (\vr) : = \sum_{i=1}^{d}\partial_{\vz^{i}} \vr^{i}(\vz)$ is the divergence operator of a vector-valued function $\vr: \sR^d \to \sR^d$, where $\vz^{i}$ and $\vr^{i}$ are the $i$ th element of $\vz$ and $\vr$; $\nabla$ is the gradient of a scalar-valued function.
$\nabla_{W_2}F(\mu_t):= div(\mu_t \nabla \frac{\delta F}{\delta \mu}(\mu_t))$ 
is the Wasserstein gradient of functional $F$ at $\mu_t$, where $\frac{\delta F}{\delta \mu}(\mu_t)$ is the first variation of $F$ at $\mu_t$.  The first variation of a \textit{functional} in function space is analogous to the first-order gradient of a \textit{function} in Euclidean distance. We put more detailed definition in Appendix \ref{sec:COV}. Intuitively, the WGF describes that the probability measure $\mu_t$ follows the steepest curve of the functional $F(\mu)$ in Wasserstein space of probability measures (function space) to gradually move towards the target probability measure.  

 We define the energy functional $F(\mu)$ for memory evolution as the following:
\begin{gather}\label{eq:func}
  F(\mu) = V(\mu) + \gD (\mu||\pi) \\
  V(\mu) = - \mathbb{E}_{\mu} \gL(\vtheta,\! \vx,\! y)\!-\!\beta \mathbb{E}_{\mu} \nabla_{\vtheta}\gL(\vtheta, \!\vx, \!y)\!\cdot\! \nabla_{\vtheta}\gL(\vtheta, \!\vx^\prime\!,\!y).
\end{gather}
 By defining such energy functional $F(\mu)$, the Eq. (\ref{eq:opt}) can be equivalently solved by the following gradient flow system Eq. (\ref{eq:gf}, \ref{eq:gf2}), we name it as \textbf{Dynamic DRO}.

 \vspace{-0.8cm}
 
\begin{numcases}{}
  \partial_t \mu_t & $= div\left(\mu_t \nabla \frac{\delta F}{\delta \mu}(\mu_t)\right)$ \label{eq:gf}; \\
  \frac{d\vtheta}{dt} & $= -\nabla_{\vtheta} \mathbb{E}_{\mu_t}\gL(\vtheta, \vx, y)$, \label{eq:gf2}
\end{numcases}

where Eq. (\ref{eq:gf}) solves the inner $\sup$ problem in Eq. (\ref{eq:opt}) with WGF in Wasserstein space and Eq. (\ref{eq:gf2}) solves the outer minimization problem in Eq. (\ref{eq:opt}) for parameter update with gradient flow in Euclidean space. In the following, we focus on solving Eq. (\ref{eq:gf}) and Eq. (\ref{eq:gf2}). 

\subsection{Training Algorithm for Dynamic DRO}
\label{sec:WGF3}

We propose three different methods for efficiently solving the Dynamic DRO in Eq.~ (\ref{eq:gf})-Eq.~(\ref{eq:gf2}). The first solution is Langevin dynamics with a diffusion process to evolve the memory data distribution; we name this method WGF-LD. Next, different from WGF-LD, which relies on \textit{randomness} to transform the memory data, we kernelize the WGF in Eq. (\ref{eq:gf}) by solving the WGF in reproducing kernel Hilbert space (RKHS). It \textit{deterministically} transforms the memory data; we name this method WGF-SVGD. Furthermore, we generalize the above WGF and improve their flexibility to incorporate prior knowledge or geometry information. One instantiation of this general WGF uses Hamiltonian dynamics; we name it WGF-HMC. More novel memory evolution methods are worth exploring in future work. 
 
Before introducing the proposed solutions, we first present some preliminaries for solving the WGF. By calculus of variation \cite{variation}, the first variation for the KL divergence and $V(\mu)$ are \cite{gradientflow}:
\begin{small}
\begin{gather*}
    \frac{\delta \gD(\mu||\pi)} {\delta \mu} =  \log \frac{\mu}{\pi} + 1; \\
    \frac{\delta V(u)}{\delta \mu}\!=\!U(\vx,\! \vtheta) \!=\! -\! \gL(\vtheta,\!\vx,\! y)\!-\!\beta \nabla_{\vtheta}\gL(\vtheta,\! \vx,\! y)\!\cdot\! \nabla_{\vtheta}\gL(\vtheta,\! \vx^\prime\!,\! y). 
\end{gather*}
\end{small}
\paragraph{Langevin Dynamics for Dynamic DRO.}
\begin{figure}
\centering
\subfloat[$t=t_0$\\]{\includegraphics[width=0.15\textwidth]{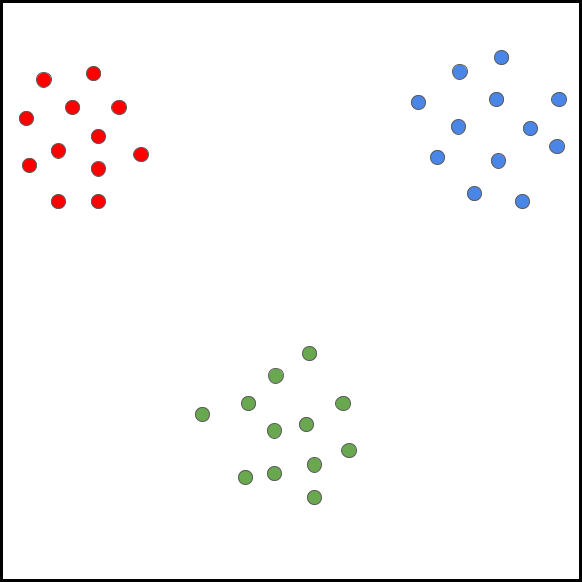}} \,
\subfloat[$t=t_1$ \\]{\includegraphics[width=0.15\textwidth]{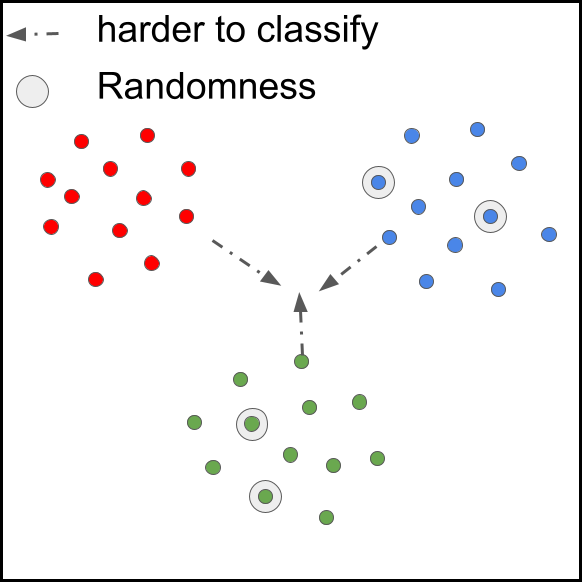}} \,
\subfloat[$t=t_2$ \\]{\includegraphics[width=0.15\textwidth]{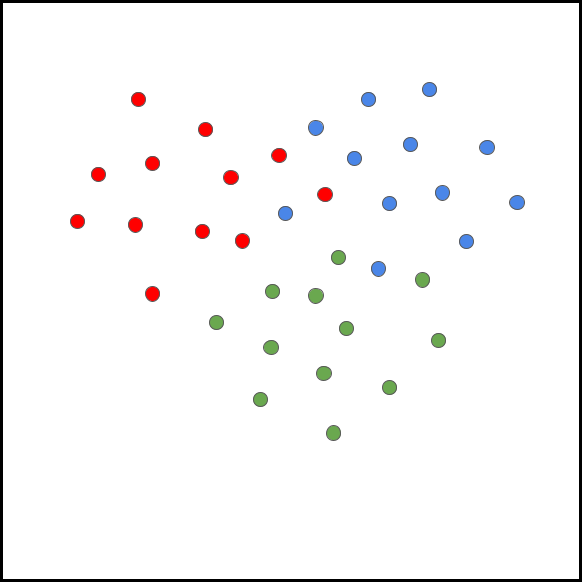}}
\caption{Illustration of WGF-LD for memory evolution at different time $t_0 < t_1 <t_2$. Initially ($t= t_0$), the raw memory data is easy to overfit. From  $t=t_1$ to $t=t_2$, the memory data becomes harder to classify and overfit. The black arrow (corresponds to the first term ($\nabla_{\vx}U(\vx_t^i, \vtheta)$)), drives the memory data to become harder to classify. The white circle (corresponds to the second term (Brownian motion)) serves as a random force such that the memory data becomes more diverse.}
\label{fig:SGLD} 
\end{figure}

If we directly use the energy functional $F(\mu)$ (Eq. (\ref{eq:func})) in Eq. (\ref{eq:gf}), solving gradient flow Eq. (\ref{eq:gf}) corresponds to the Langevin dynamics with the following stochastic differential equation \cite{SGLD11}:
\begin{align}
       dX = -\nabla_{X}U(X, \vtheta) dt + \sqrt{2} dW_t, \label{eq:SGLD}
\end{align}
where  $X = (X_t)_{t\geq 0}$ is the memory evolution process as previously defined. $W = (W_t)_{t\geq 0}$ is the standard Brownian motion in $\sR^{n}$ \cite{bass_2011}. If $X_t \sim \mu_t$ evolves according to the Langevin dynamics Eq. (\ref{eq:SGLD}) in Euclidean space, then $\mu(\vx, t) = \mu_t(\vx)$ evolves according to gradient flow Eq. (\ref{eq:gf}) in the space of probability measures \cite{SGLDWGF}. If we discretize the above equation and view each datapoint in memory as one particle, the memory buffer data evolves with Langevin dynamics to
obtain diverse memory data by the following updates:

\begin{equation} \label{eq:particleSGLD}
    \vx_{t+1}^{i} - \vx_t^{i} = -\alpha(\nabla_{\vx} U(\vx_t^{i}, \vtheta)) + \sqrt{2\alpha} \xi_t.
\end{equation}
As illustrated in Figure \ref{fig:SGLD}, the first term in Eq. (\ref{eq:particleSGLD}) drives the particles (memory data) towards the worst-case memory data distribution $\pi$ to make the memory data gradually harder to be memorized by dynamically increasing the energy functional. For the second term, it adds noise to encourage diversity in the transformed memory data, where $\xi_t$ is standard Gaussian noise, and $\alpha$ is step size, i.e., a proper amount of added Gaussian noise tailored to the used step size. We name this memory evolution method as \textbf{WGF-LD}. 

\paragraph{Kernelized Method for Dynamic DRO.}

\begin{figure}
\centering
\subfloat[$t=t_0$\\]{\includegraphics[width=0.15\textwidth]{visnew1.png}} \,
\subfloat[$t=t_1$ \\]{\includegraphics[width=0.15\textwidth]{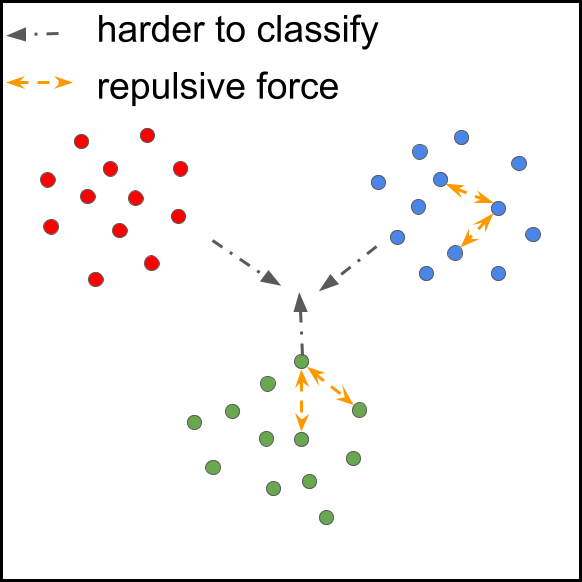}} \,
\subfloat[$t=t_2$ \\]{\includegraphics[width=0.15\textwidth]{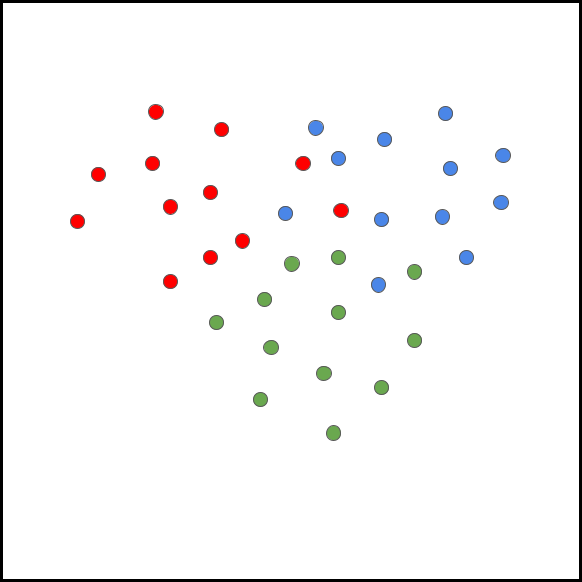}}
\caption{Illustration of WGF-SVGD for memory evolution at different time $t_0 < t_1 <t_2$. Initially ($t= t_0$), the raw memory data is easy to overfit. From $t=t_1$ to $t=t_2$, the memory data becomes harder to classify and overfit. The black arrow (corresponds to the first term ($k(\vx_t^i, \vx_t^j) \nabla_{\vx_t^j} U(\vx_t^j, \vtheta)$))  drives the memory data to become harder to classify. The orange arrow (corresponds to the second term ($\nabla_{\vx_t^j} k(\vx_t^i, \vx_t^j)$)) serves as repulsive force such that the memory data becomes more diverse.}
\label{fig:SVGD} 
\end{figure} 

We replace the Wasserstein gradient $\nabla_{W_2}F(\mu_t)$  
by the transformation $\gK_{\mu} \nabla_{W_2}F(\mu_t)$ under the integral operator $\gK_{\mu}f(\vx) = \int K(\vx, \vx^{\prime})f(\vx^{\prime})d\mu(\vx^{\prime})$; where the RKHS space induced by the kernel $K$ is denoted by $\gH$. 
The probability measure in the kernelized Wasserstein space actually follows the kernelized WGF \cite{liu2017stein}:
\begin{equation} \label{eq:kernelWGF}
     \partial_t \mu_t = div(\mu_t \gK_{\mu_t} \nabla \frac{\delta F}{\delta \mu}(\mu_t)).
\end{equation}
Eq. (\ref{eq:kernelWGF}) can be viewed as the WGF Eq. (\ref{eq:gf}) in RKHS. It indicates that the random variable $X_t$ which describes the evolved memory data at time $t$ evolves as the following differential equation \cite{liu2017stein}:
\begin{equation}
   \frac{d X}{dt} = - [\gK_{\mu} \nabla \frac{\delta F}{\delta \mu}(\mu_t)](X).
\end{equation}

This kernelized version is the deterministic approximation of the WGF in Eq. (\ref{eq:gf}) \cite{liu2019stein}. If we discretize the above equation and view each datapoint in memory as one particle, we can obtain the following memory evolution update equation:
\begin{equation}
\small \label{eq:particleSVGD}
    \vx_{t+1}^i -  \vx_{t}^i  = -\frac{\alpha}{N} \sum_{j=1}^{j= N}  [\underbrace{k(\vx_t^i, \vx_t^j) \nabla_{\vx_t^j} U(\vx_t^j, \vtheta)}_{\text{smoothed gradient}} +  \underbrace{\nabla_{\vx_t^j} k(\vx_t^i, \vx_t^j)}_{\text{repulsive term}}],
\end{equation}
As illustrated in Figure \ref{fig:SVGD}, the first term drives the memory data towards the worst-case memory data distribution by increasing the energy functional. The update is driven by the kernel weighted sum of the gradients from the memory data points, thus smoothing the memory data gradients. The second term serves as a repulsive force that prevents the memory data points from collapsing into a single mode, thus diversifying the memory data population. In this paper, we use Gaussian kernel $k(\vx_i, \vx_j) = exp (- \frac{(\vx_i - \vx_j)^2}{2\sigma^2})$. We name this memory evolution method as \textbf{WGF-SVGD}. We put detailed derivations of this evolution equation in Appendix \ref{sec:derivationsevolution}.

\paragraph{General Memory Evolution for Dynamic DRO.} \cite{ma2015complete} found that any continuous Markov process that provides samples from the target distribution
can be written in a very general sampler form. The corresponding general WGF for memory evolution can be written as:
\begin{equation}
       \partial_t \mu_t = div(\mu_t (\mD + \mQ) \nabla \frac{\delta F}{\delta \mu}(\mu_t)),
\end{equation}
Where $\mD$ is a positive semidefinite diffusion matrix, $\mQ$ is a skew-symmetric curl matrix representing the deterministic traversing effect \cite{ma2015complete}. One particular case is:
$$\mD = \begin{pmatrix}
  0 & 0\\ 
  0 & \mC
\end{pmatrix}, 
\mQ = \begin{pmatrix}
  0 & -\mI\\ 
  \mI & 0
\end{pmatrix},$$ 
Where $\mC$ is the friction term, and $\mI$ is the identity matrix. This WGF corresponds to Hamiltonian dynamics \cite{pmlr-v32-cheni14} and can be solved by the following evolution:
\begin{equation}
\left\{
\begin{aligned}  \label{eq:HMC}
  \vx_{t+1} -   \vx_{t} & = \vv_t,\\
  \vv_{t+1} -   \vv_{t} & = - \alpha \nabla U(\vx, \vtheta) -\tau \vv + \sqrt{2\tau \alpha} \xi_t,\\
\end{aligned}
\right.
\end{equation}
Where $\vv$ is the momentum variable, and $\tau$ is the momentum weight. We name this method as \textbf{WGF-HMC}.

The most attractive property of this WGF for memory distribution evolution is that we can freely specify the matrix $\mQ$ and $\mD$  tailored to practical requirements. We can consider prior knowledge or geometry information by designing tailored $\mQ$ and $\mD$, or develop the kernelized version of this general WGF. These research directions specialized for CL are left as future work to explore. Our framework is quite flexible and general due to the energy functional design and WGF specification.

The proposed memory evolution algorithm is shown in Algorithm \ref{alg:DROtrain}, with the flexibility to use various evolution methods. Line 3-4 describes that a mini-batch data arrives at time $k$ and samples a mini-batch data from the memory buffer. Line 5-7 is to evolve the mini-batch memory data with $T$ steps depending on using which evolution methods. Line 8 updates the model parameters with the evolved memory data and mini-batch data received at time $k$. Line 9 updates the memory buffer with the mini-batch data received at time $k$ using reservoir sampling. 
Note that we do not replace the raw memory data with the evolved memory data for implementation simplicity in the current version. If we replace the raw memory data with the evolved one, we can reduce the number of evolution steps needed at each CL timestamp to improve efficiency since an example can be adequately evolved after accumulating many timestamps. Earlier examples can be evolved less frequently than the later ones. But replacing the raw memory data with the evolved one could slightly reduce the performance in our experiment. Our method needs to store a mini-batch of evolved memory data. This memory cost is negligible compared to the entire data stream memory buffer. We can view memory evolution from two perspectives. First, \textit{local evolution} evolves the current raw memory data distribution with several adaptation steps. Second, \textit{global evolution} evolves the memory data at different CL timestamps due to different model parameters. 

\textbf{Discrete input.} For the application of our methods in the discrete input domain (such as text), text can be embedded in continuous space (e.g., word embedding) and then evolve on embedding with our methods.

  \begin{algorithm}[h]
  \small
	\caption{\small Distributionally Robust Memory Evolution.}
	\label{alg:DROtrain}
	\begin{algorithmic}[1]
		\STATE{\bf REQUIRE: } model parameters $\vtheta$, learning rate $\eta$, evolution rate (step size) $\alpha$, number of evolution steps $T$ at each iteration,  memory buffer $\gM$; $K$ is the number of mini-batch data in the data stream. 
		 \vspace{-0.0cm}
	    	    \FOR{$k = 1$ to $K$}
	    	    \STATE a new mini-batch data $(\vx_k, y_k)$  arrives. 
	    	    \STATE sample mini-batch from memory buffer, i.e., $(\vx, y) \sim \gM$
	    	    \FOR{$t = 1$ to $T$}
                \STATE $(\vx, y) = Evolve((\vx, y))$ by WGF-LD (Eq. (\ref{eq:particleSGLD})) or WGF-SVGD (Eq. (\ref{eq:particleSVGD}))) or WGF-HMC (Eq.  (\ref{eq:HMC})). 
                \ENDFOR
                   \STATE $\vtheta_{k+1} = \vtheta_{k} - \eta \nabla_{\vtheta} [ \gL( \vtheta_{k}, \vx, y) + \gL(\vtheta_k, \vx_k, y_k)$]
                \STATE update memory buffer by reservoir sampling,   
                $\gM = \mbox{reservoirsampling} (\gM, (\vx_k, y_k))$
                \ENDFOR
                 \vspace{-0.1cm}
	\end{algorithmic}
\end{algorithm}

%% file: experiment.tex
\section{Experiments}
To evaluate the effectiveness of our memory evolution methods, we compared a variety of state-of-the-art baseline approaches. We first describe the benchmark datasets and baselines in Section \ref{exp-setup}. Then, we compare various baselines on several datasets in Section \ref{sec:CLresults}, and evaluate the methods in terms of robustness to adversarial examples in Section \ref{sec:attack}. We perform ablation study in Section \ref{sec:ablation}.

\subsection{Experiment Setup}\label{exp-setup}
\textbf{CIFAR10}, following \cite{NIPS2019_9357}, we split the CIFAR-10 dataset into 5 disjoint tasks with the same training, validation, and test sets. 

\textbf{MiniImagenet}, following \cite{NIPS2019_9357}, we split MiniImagenet \cite{NIPS2016_90e13578} dataset with 100 image classes into 20 disjoint tasks. Each task has 5 classes.

\textbf{CIFAR-100} \cite{cifar100}, contains 100 image classes. We also split it into 20 disjoint tasks.

\textbf{Baselines.} We performed comprehensive experiments by comparing to the following strong baselines, including \textit{Experience Replay (ER) \cite{ERRing19}}, \textit{Maximally Interfering
Retrieval (MIR) \cite{NIPS2019_9357}}, \textit{AGEM \cite{AGEM19}}, Gradient-Based Sample Selection (GSS-Greedy) \cite{aljundi2019gradient} and GMED \cite{jin2021gradientbased}. Furthermore,  following \cite{jin2021gradientbased}, we also compare data augmentation, such as random rotations, scaling, and horizontal
flipping applied to memory buffer data in ER and name this baseline as ER$_{aug}$. Following \cite{NIPS2019_9357}, we also compare (1) fine-tuning, which trains on each latent task sequentially when new batches of each task arrive without any forgetting mitigation mechanism; (2) iid online: which trains the model with a single-pass through the iid sampled data
on the same set of samples; (3) iid offline: (upper-bound) which trains the model with multiple passes through the iid sampled data. We train the model with 5 epochs for this baseline. Detailed descriptions of the compared baselines are placed in Appendix \ref{sec:baseline}. 

In addition, our proposed methods are orthogonal to existing memory-replay-based CL methods. Thus, they are versatile, and we can seamlessly combine the proposed methods with them. For illustration, we combine our proposed methods with ER, MIR, and GMED to show the effectiveness. We name the combination methods ER+WGF-LD,  ER+WGF-SVGD, ER+WGF-HMC, MIR+WGF-LD,  GMED+WGF-LD, etc. Combining with other memory-replay-based methods is straightforward. 

\textbf{Implementation Details.}
We use the Resnet-18 as \cite{NIPS2019_9357}. The number of evolution steps $T$ is set to be $5$, the evolution rate $\alpha$ is set to be $0.01$ for CIFAR10, $0.05$ for CIFAR100 and $0.001$ for MiniImagenet, $\beta =0.003$ and momentum $\tau = 0.1$. We set the memory buffer size to be 500 for CIFAR-10 dataset,  $5K$ for CIFAR-100 and $10K$ for MiniImagenet. All other hyperparameters are the same as \cite{NIPS2019_9357}. All reported results in our experiments are the average accuracy of 10 runs with standard deviation.

\subsection{Comparison to Continual Learning} \label{sec:CLresults}

We compare the proposed methods to various task-free CL baselines. Table \ref{tab:baseline} shows the effectiveness of combining the proposed method with existing memory-replay methods, e.g., ER, MIR and GMED. We can observe that our method outperforms these strong baselines. In particular, for ER and ER + WGF methods, our method outperforms baselines by $4.5\%$, $1.4\%$, and $2.8\%$ on CIFAR10, CIFAR-100, and MiniImageNet, respectively. For  MIR and MIR + WGF methods, our method outperforms baselines by $3.8\%$, $1.6\%$, and $2.1\%$ on CIFAR10, CIFAR-100, and MiniImageNet, respectively. For GMED and GMED + WGF methods, our method outperforms baselines by $3.6\%$, $0.9\%$, and $1.4\%$ on CIFAR10, CIFAR-100, and MiniImageNet, respectively. 
Our methods outperform baselines because they dynamically evolve the memory data distribution and sufficiently explore the input space in a principled way. The proposed methods generate more diverse memory data, and the evolved memory becomes more difficult for the CL model to memorize than baseline methods. In addition, WGF-SVGD generally performs better than WGF-SGLD and WGF-HMC. We believe this is because, in RKHS, the evolved memory data is encouraged to be far apart by the kernel repulsive forces; the evolved memory data may better represent the distribution of all the previous data.

\begin{table}
\centering
\caption{Comparison to continual learning baselines on CIFAR10, CIFAR-100 and MiniImagenet by combing our proposed method with existing CL methods}
\begin{adjustbox}{scale=0.82,tabular= lccc,center}
\begin{tabular}{lrrrrrrr|c|}
\toprule
\textbf{Algorithm} & \textbf{CIFAR10} & \textbf{CIFAR-100} &\!\!\!\textbf{MiniImagenet}\!\!\!\\
\midrule
fine-tuning &$18.9\pm 0.1$ & $3.1\pm 0.2$ & $2.9 \pm 0.5$\\
A-GEM &$19.0\pm 0.3$ & $2.4\pm 0.2$  & $3.0 \pm 0.4$\\
GSS-Greedy &$29.9 \pm 1.5$ & $19.5\pm 1.3$& $17.4 \pm 0.9$ \\
\midrule
ER & $33.3\pm 2.8$ & $20.1\pm 1.2$ & $24.8 \pm 1.0$ \\
ER + WGF-LD &$37.6\pm 1.5$ &  \textbf{21.5 $\pm$ 1.3} & $27.3 \pm 1.0$  \\
ER + WGF-SVGD\!\!\! &$36.5\pm 1.4$ &$21.3\pm 1.5$  & \textbf{27.6  $\pm$ 1.3}\\
ER + WGF-HMC &\textbf{37.8 $\pm$ 1.3} &$21.2\pm 1.4$ & $27.2 \pm 1.1$\\
\midrule
MIR &$34.4\pm 2.5$ & $20.0\pm 1.7$& $25.3\pm 1.7$\\
MIR + WGF-LD &\textbf{38.2 $\pm$ 1.2}& \textbf{21.6 $\pm$ 1.2}  & $26.9\pm 1.0$\\
MIR + WGF-SVGD\!\!\! & $37.0\pm 1.4$ &$21.2\pm 1.5$  & \textbf{27.4 $\pm$ 1.2}\\
MIR + WGF-HMC & $37.9\pm 1.5$ &$21.3\pm 1.4$  & $27.1\pm 1.3$\\
\midrule
GMED (ER) & $34.8 \pm 2.2$ & $20.9\pm 1.6$ & $27.3\pm 1.8$\\
GMED + WGF-LD & \textbf{38.4 $\pm$ 1.6} &$21.7\pm 1.7$ &  $28.3\pm 1.9$\\
GMED + WGF-SVGD\!\!\!\! & $37.6\pm 1.7$ & \textbf{21.8 $\pm$ 1.5}  &  \textbf{28.7 $\pm$ 1.5}\\
GMED + WGF-HMC & $37.8\pm 1.2$ &$21.5\pm 1.9$ &  $28.4\pm 1.3$\\
\midrule
ER$_{aug}$ + ER & $46.3\pm 2.7$ & $18.3\pm 1.9$ & $30.8\pm 2.2$\\
ER$_{aug}$ + WGF-LD & $47.6\pm 2.4$   &$19.8\pm 2.2$  & $31.9\pm 1.8$\\
ER$_{aug}$ + WGF-SVGD\!\!\! & \textbf{47.9 $\pm$ 2.5}  &$19.9\pm 2.3$ & \textbf{32.2 $\pm$ 1.5}\\
ER$_{aug}$ + WGF-HMC &$47.8\pm 2.6$ & \textbf{20.3 $\pm$ 2.1}  &  $31.7\pm 2.0$ \\
\midrule
iid online &$60.3\pm 1.4$  & $18.7\pm 1.2$ & $17.7\pm 1.5$\\
iid offline & $78.7\pm 1.1$  &$44.9\pm 1.5$ & $39.8\pm 1.4$ \\
\bottomrule
\end{tabular}
\label{tab:baseline}
\end{adjustbox}
\vspace{-0.4cm}
\end{table}

\begin{table}[htb]
\centering
\vspace{-0.3cm}
\caption{Carlini $\&$ Wagner attack model performance for the CIFAR-100 and Mini-Imagenet dataset}
\begin{adjustbox}{scale=0.85,tabular= lccc,center}
\begin{tabular}{lrrrrrrr|c|}
\toprule
\textbf{Algorithm}& CIFAR-10 & CIFAR-100 & Mini-Imagenet\\
\midrule
ER & $2.0\pm 0.1$ & $0.0$ &  $0.0$  \\
GMED &$2.1\pm 0.1$ &  $0.0$ &  $0.0$  \\
ER + WGF-LD & $8.0\pm 0.2$ &  \textbf{3.0 $\pm$ 0.2} &  \textbf{3.1 $\pm$ 0.1}   \\
ER + WGF-SVGD &$4.2\pm 0.1$ &  $0.0$ &  $2.2\pm 0.2$     \\
ER + WGF-HMC & \textbf{8.2 $\pm$ 0.3}  &  $2.5\pm 0.2$  &  $3.0\pm0.1$  \\
\bottomrule
\end{tabular}
\label{tab:CWattack}
\end{adjustbox}
\vspace{-0.4cm}
\end{table}

\subsection{Robustness to Adversarial Perturbations} \label{sec:attack}
In this section, we evaluate the robustness of the CL model to adversarial perturbed examples. Given a classifier $f(\vx, \vtheta)$, for an image $\vx$, the goal is to find another example  $\vx^{\prime}$ that is close enough to $\vx$ measured by some distance function $D(\vx. \vx^{\prime}) \leq \epsilon$ such that the classifier classifies it into another different class, i.e. $f(\vx, \vtheta) \neq f(\vx^{\prime}, \vtheta)$. In this paper, we focus on the most commonly used  $\ell_{\infty}$ and $\ell_{2}$ norm as the distance function. 

\begin{figure}[h]
\centering
\includegraphics[width=0.5\textwidth]{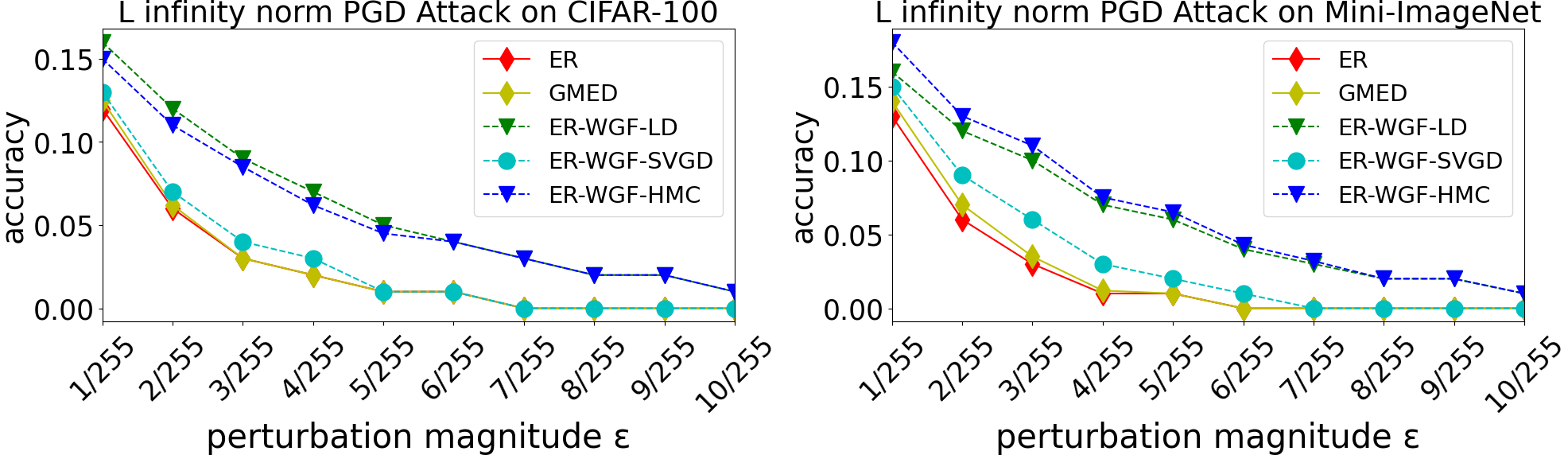}
\vspace{-0.7cm}
\caption{PGD $\ell_{\infty}$ attack results on two datasets: CIFAR-100 (left) and Mini-ImageNet (right).}  
\vspace{-0.4cm}
\label{fig:PGD}
\end{figure}

Our memory evolution methods optimize on the worst-case evolved memory data distribution during CL; the model would be thus naturally robust to adversarial perturbations. For $\ell_{\infty}$ norm attack, we evaluate the robustness under the strong PGD $\ell_{\infty}$ attack \cite{madry2019deep}, which constructs adversarial examples with projected gradient descent and $\ell_{\infty}$ norm constraints. The adversarial perturbation magnitude ranges from $[1/255, 2/255, \cdots, 10/255]$ with 20 steps attack and stepsize of $\frac{2}{255}$ on CIFAR-100 and Mini-ImageNet. For $\ell_{2}$ norm attack, we adopt the strong Carlini $\&$ Wagner attack \cite{carlini2017evaluating}. For illustration, we evaluate the model robustness to adversarial examples after training with ER, GMED, ER + WGF-LD, ER + WGF-SVGD, and ER + WGF-HMC, respectively. Figure \ref{fig:PGD} shows the PGD $\ell_{\infty}$ attack results on CIFAR-100 and Mini-ImageNet. Our WGF-HMC and WGF-LD memory evolution significantly outperforms naive ER baseline by $4\%-12\%$ depending on the perturbation magnitude. In addition, WGF-HMC and WGF-LD perform more robust than WGF-SVGD. We believe this is because the randomness introduced in WGF-HMC and WGF-LD can better explore the input space and thus can generate harder evolved memory data. WGF-SVGD smooths the function gradient by the kernel function, thus may generate less hard examples. Table \ref{tab:CWattack} shows the  Carlini $\&$ Wagner attack results. We can see that under the strong Carlini $\&$ Wagner attack, ER baseline accuracy becomes zero, and our methods still outperform baselines ranging from $6.1\%, 3.0\%, 3.1\%$ on CIFAR-10, CIFAR-100, Mini-ImageNet, respectively. Both results demonstrate the robustness of our proposed methods to adversarial examples.

\subsection{Ablation Study}
\label{sec:ablation}
\textbf{Effects of different memory size}. 
To investigate the effect of different smaller memory sizes on the model performance, we evaluate the effects on Mini-ImageNet with memory sizes of 3000, 5000, and 10000. We evaluate the effects on CIFAR-100 with the memory sizes of 2000, 3000, and 5000. We show the results in Table \ref{tab:memory_CIFARFS_Mini}. In most cases, our WGF memory evolution substantially outperforms the baselines with different memory buffer sizes.

\begin{table}[htb]
\centering
\caption{Effect of different memory size on the model performance for the CIFAR-100 and Mini-Imagenet datasets.}
\begin{adjustbox}{scale=0.85,tabular= lccc,center}
\begin{tabular}{lrrrrrrr|c|}
\toprule
\multicolumn{5}{c}{CIFAR-100} \\
\hline
\textbf{Memory Size}& & 2000 & 3000 & 5000\\
\midrule
ER &&$11.2\pm 1.0$  &$15.0\pm 0.9$ & $20.1\pm1.2$\\
ER + WGF-LD && \textbf{12.9 $\pm$ 1.2} &$17.0\pm 1.1$ &\textbf{21.5 $\pm$ 1.3} \\
ER + WGF-SVGD &&$12.3\pm 1.1$  &$16.0\pm 1.2$  & $21.3\pm1.5$\\
ER + WGF-HMC &&$12.7\pm 1.0$ &\textbf{17.2 $\pm$ 1.0}   &$21.2\pm1.4$ \\
\midrule
MIR &&$11.6\pm 0.8$ &$15.6\pm 1.0$ &  $20.0\pm1.7$\\
MIR + WGF-LD &&$13.1\pm 0.9$ &$17.3\pm 1.2$  & \textbf{21.6 $\pm$ 1.2} \\
MIR + WGF-SVGD &&$12.7\pm 1.0$ &$16.5\pm 1.3$  &$21.2\pm 1.5$  \\
MIR + WGF-HMC &&\textbf{13.2 $\pm$ 1.2} & \textbf{17.5 $\pm$ 1.1} &$21.3\pm 1.4$  \\
\bottomrule
\toprule
\multicolumn{5}{c}{Mini-Imagenet} \\
\hline
\textbf{Memory Size}& & 3000 & 5000 & 10000\\
\midrule
ER && $13.4\pm 1.4$ &  $17.9\pm 1.6$&  $24.8\pm 0.9$\\
ER + WGF-LD &&\textbf{16.2 $\pm$ 1.2} & $20.8\pm 1.2$ &$27.3\pm 1.0$\\
ER + WGF-SVGD &&$15.7\pm 1.2$ & \textbf{21.3 $\pm$ 1.0} & \textbf{27.6 $\pm$ 1.3} \\
ER + WGF-HMC &&$15.9\pm 1.5$ & $20.6\pm 1.4$ &$27.2\pm 1.1$\\
\midrule
MIR && $12.6\pm 1.5$ & $17.4\pm 1.2$&  $25.3\pm 1.7$\\
MIR + WGF-LD && $15.5\pm 1.4$  & $20.5\pm 1.1$&  $26.9\pm 1.0$\\
MIR + WGF-SVGD &&$15.3\pm 1.2$  & \textbf{20.7 $\pm$ 1.6}&  \textbf{27.4 $\pm$ 1.2}\\
MIR + WGF-HMC && \textbf{15.8 $\pm$ 1.7}  & $20.3\pm 1.5$&  $27.1\pm 1.3$ \\
\bottomrule
\end{tabular}
\label{tab:memory_CIFARFS_Mini}
\end{adjustbox}
\end{table}

\textbf{Effect of number of evolution steps}.
To investigate the effect of different evolution steps, we compare 3, 5, and 7 evolution steps, respectively. Table \ref{tab:evolution_steps} shows the performance variation of different number of evolution steps. We can find that the performance improves slightly with an increasing number of evolution steps. For efficiency and sufficiently exploring the input space to evolve harder memory examples, we choose the evolution step of 5.

\begin{table}[htb]
\vspace{-0.3cm}
\centering
\caption{Effect of number of evolution steps on Mini-ImageNet.}
\begin{adjustbox}{scale=0.87,tabular= lccc,center}
\begin{tabular}{lrrrrrrr|c|}
\toprule
\textbf{Evolution Steps}& 3 & 5 & 7\\
\midrule
ER + WGF-LD &$27.0 \pm 0.9$ &  $27.3 \pm 1.0$  & $27.5\pm 1.4$\\
ER + WGF-SVGD &$27.2 \pm 1.2$  & 27.6  $\pm$ 1.3 &$27.2 \pm 1.2$\\
ER + WGF-HMC &$27.1 \pm 1.3$ & $27.2 \pm 1.1$& $27.6 \pm 1.0$ \\
\bottomrule
\end{tabular}
\label{tab:evolution_steps}
\end{adjustbox}
\vspace{-0.1cm}
\end{table}

\textbf{Hyperparameter sensitivity}.
Due to space limitation, we put hyperparameter sensitivity analysis, including the regularizer weight $\beta$ and evolution rate $\alpha$, in Appendix \ref{sec:appresult}.

\textbf{Computation cost}.
    We compare the proposed ER + WGF to ER to evaluate its running efficiency. Table \ref{tab:efficiency} shows the efficiency comparison results. We set the simple baseline ER with a running time unit of 1. Our method has 3-4 times the computational cost compared to a simple ER baseline. In future work, we will improve its computational efficiency. 

\begin{table}[H]
\vspace{-0.4cm}
\centering
\caption{Computation efficiency (relative training time) of the proposed method compared to baseline.}
\begin{adjustbox}{scale=0.85,tabular= lccc,center}
\begin{tabular}{lrrrrrrr|c|}
\toprule
\textbf{Algorithm}&  running time \\
\midrule
ER & 1.0\\
ER + WGF-LD & 3.4 \\
ER + WGF-SVGD & 4.1 \\
ER + WGF-HMC & 3.5 \\
\bottomrule
\end{tabular}
\label{tab:efficiency}
\end{adjustbox}
\vspace{-0.4cm}
\end{table}

%% file: conclusion.tex
\section{Conclusion}
This paper proposes a novel concept of DRO memory evolution for task-free continual learning to dynamically evolve the memory data distribution to mitigate the memory overfitting issue and fill the gap between the memory data distribution and the distribution of all the previous data. We propose a family of memory evolution methods with WGF. The proposed principled framework is general, flexible, and easily expandable. 
 Future work includes designing more informative functional and novel gradient flow dynamics to incorporate physical intuitions and geometry constraints. Extensive experiments compared to various state-of-the-art methods demonstrate the effectiveness of the proposed method. Interestingly, our methods are more robust to adversarial examples than compared baselines. 
 
 \section{Acknowledgement}
 
 We thank all the anonymous reviewers for their insightful and thoughtful comments. This research was supported in part by NSF through grant IIS-1910492.

%% file: main_paper.bbl
\begin{thebibliography}{40}
\providecommand{\natexlab}[1]{#1}
\providecommand{\url}[1]{\texttt{#1}}
\expandafter\ifx\csname urlstyle\endcsname\relax
  \providecommand{\doi}[1]{doi: #1}\else
  \providecommand{\doi}{doi: \begingroup \urlstyle{rm}\Url}\fi

\bibitem[Aljundi et~al.(2019{\natexlab{a}})Aljundi, Belilovsky, Tuytelaars,
  Charlin, Caccia, Lin, and Page-Caccia]{NIPS2019_9357}
Aljundi, R., Belilovsky, E., Tuytelaars, T., Charlin, L., Caccia, M., Lin, M.,
  and Page-Caccia, L.
\newblock Online continual learning with maximal interfered retrieval.
\newblock \emph{Advances in Neural Information Processing Systems 32}, pp.\
  11849--11860, 2019{\natexlab{a}}.

\bibitem[Aljundi et~al.(2019{\natexlab{b}})Aljundi, Kelchtermans, and
  Tuytelaars]{aljundi2019taskfree}
Aljundi, R., Kelchtermans, K., and Tuytelaars, T.
\newblock Task-free continual learning.
\newblock \emph{Proceedings of the IEEE Conference on Computer Vision and
  Pattern Recognition (CVPR)}, 2019{\natexlab{b}}.

\bibitem[Aljundi et~al.(2019{\natexlab{c}})Aljundi, Lin, Goujaud, and
  Bengio]{aljundi2019gradient}
Aljundi, R., Lin, M., Goujaud, B., and Bengio, Y.
\newblock Gradient based sample selection for online continual learning.
\newblock In \emph{Advances in Neural Information Processing Systems 30},
  2019{\natexlab{c}}.

\bibitem[Ambrosio et~al.(2008)Ambrosio, Gigli, and Savare]{gradientflow}
Ambrosio, L., Gigli, N., and Savare, G.
\newblock Gradient flows: In metric spaces and in the space of probability
  measures.
\newblock \emph{(Lectures in Mathematics. ETH)}, 2008.

\bibitem[Bass(2011)]{bass_2011}
Bass, R.~F.
\newblock \emph{Stochastic Processes}.
\newblock Cambridge Series in Statistical and Probabilistic Mathematics.
  Cambridge University Press, 2011.
\newblock \doi{10.1017/CBO9780511997044}.

\bibitem[Boyd \& Vandenberghe(2004)Boyd and Vandenberghe]{boyd2004convex}
Boyd, S. and Vandenberghe, L.
\newblock \emph{Convex optimization}.
\newblock Cambridge university press, 2004.

\bibitem[Carlini \& Wagner(2017)Carlini and Wagner]{carlini2017evaluating}
Carlini, N. and Wagner, D.
\newblock Towards evaluating the robustness of neural networks.
\newblock \emph{2017 IEEE Symposium on Security and Privacy (SP)}, 2017.

\bibitem[Chaudhry et~al.(2019{\natexlab{a}})Chaudhry, Ranzato, Rohrbach, and
  Elhoseiny]{AGEM19}
Chaudhry, A., Ranzato, M., Rohrbach, M., and Elhoseiny, M.
\newblock Efficient lifelong learning with a-gem.
\newblock \emph{Proceedings of the International Conference on Learning
  Representations}, 2019{\natexlab{a}}.

\bibitem[Chaudhry et~al.(2019{\natexlab{b}})Chaudhry, Rohrbach, Elhoseiny,
  Ajanthan, Dokania, Torr, and Ranzato]{ERRing19}
Chaudhry, A., Rohrbach, M., Elhoseiny, M., Ajanthan, T., Dokania, P.~K., Torr,
  P. H.~S., and Ranzato, M.
\newblock Continual learning with tiny episodic memories.
\newblock \emph{https://arxiv.org/abs/1902.10486}, 2019{\natexlab{b}}.

\bibitem[Chen et~al.(2014)Chen, Fox, and Guestrin]{pmlr-v32-cheni14}
Chen, T., Fox, E., and Guestrin, C.
\newblock Stochastic gradient hamiltonian monte carlo.
\newblock In \emph{Proceedings of the 31st International Conference on Machine
  Learning}, 2014.

\bibitem[Chewi et~al.(2020)Chewi, Le~Gouic, Lu, Maunu, and
  Rigollet]{NEURIPS2020_SVGD}
Chewi, S., Le~Gouic, T., Lu, C., Maunu, T., and Rigollet, P.
\newblock Svgd as a kernelized wasserstein gradient flow of the chi-squared
  divergence.
\newblock In \emph{Advances in Neural Information Processing Systems}, 2020.

\bibitem[Chrysakis \& Moens(2020)Chrysakis and Moens]{pmlr-v119-chrysakis20a}
Chrysakis, A. and Moens, M.-F.
\newblock Online continual learning from imbalanced data.
\newblock \emph{Proceedings of the 37th International Conference on Machine
  Learning}, 119:\penalty0 1952--1961, 2020.

\bibitem[He et~al.(2019)He, Sygnowski, Galashov, Rusu, Teh, and
  Pascanu]{he2019task}
He, X., Sygnowski, J., Galashov, A., Rusu, A.~A., Teh, Y.~W., and Pascanu, R.
\newblock Task agnostic continual learning via meta learning.
\newblock \emph{https://arxiv.org/abs/1906.05201}, 2019.

\bibitem[Jin et~al.(2021)Jin, Sadhu, Du, and Ren]{jin2021gradientbased}
Jin, X., Sadhu, A., Du, J., and Ren, X.
\newblock Gradient-based editing of memory examples for online task-free
  continual learning.
\newblock \emph{Advances in Neural Information Processing Systems}, 2021.

\bibitem[Jordan et~al.(1998)Jordan, Kinderlehrer, , and Otto.]{SGLDWGF}
Jordan, R., Kinderlehrer, D., , and Otto., F.
\newblock The variational formulation of the fokker–planck equation.
\newblock \emph{SIAM Journal on Mathematical Analysis}, 1998.

\bibitem[Kirkpatrick et~al.(2017)Kirkpatrick, Pascanu, Rabinowitz, Veness,
  Desjardins, Rusu, Milan, Quan, Ramalho, Grabska-Barwinska, Hassabis, Clopath,
  Kumaran, and Hadsell]{EWC16}
Kirkpatrick, J., Pascanu, R., Rabinowitz, N., Veness, J., Desjardins, G., Rusu,
  A.~A., Milan, K., Quan, J., Ramalho, T., Grabska-Barwinska, A., Hassabis, D.,
  Clopath, C., Kumaran, D., and Hadsell, R.
\newblock Overcoming catastrophic forgetting in neural networks.
\newblock \emph{Proceedings of the national academy of sciences}, 2017.

\bibitem[Krizhevsky(2009)]{cifar100}
Krizhevsky, A.
\newblock Learning multiple layers of features from tiny images.
\newblock \emph{Technical report}, 2009.

\bibitem[Lee et~al.(2020)Lee, Ha, Zhang, and Kim]{lee2020neural}
Lee, S., Ha, J., Zhang, D., and Kim, G.
\newblock A neural dirichlet process mixture model for task-free continual
  learning.
\newblock In \emph{Proceedings of the 17th International Conference on Machine
  Learning}, 2020.

\bibitem[Liu et~al.(2019)Liu, Zhuo, and Zhu]{MCMCflow}
Liu, C., Zhuo, J., and Zhu, J.
\newblock Understanding mcmc dynamics as flows on the wasserstein spac.
\newblock \emph{Proceedings of the International Conference on Machine
  Learning}, 2019.

\bibitem[Liu(2017)]{liu2017stein}
Liu, Q.
\newblock Stein variational gradient descent as gradient flow.
\newblock \emph{Advances in Neural Information Processing Systems}, 2017.

\bibitem[Liu \& Wang(2016)Liu and Wang]{liu2019stein}
Liu, Q. and Wang, D.
\newblock Stein variational gradient descent: A general purpose bayesian
  inference algorithm.
\newblock \emph{Advances in Neural Information Processing Systems}, 2016.

\bibitem[Lopez-Paz \& Ranzato(2017)Lopez-Paz and Ranzato]{lopezpaz2017gradient}
Lopez-Paz, D. and Ranzato, M.
\newblock Gradient episodic memory for continual learning.
\newblock \emph{Advances in Neural Information Processing Systems}, 2017.

\bibitem[Ma et~al.(2015)Ma, Chen, and Fox]{ma2015complete}
Ma, Y.-A., Chen, T., and Fox, E.~B.
\newblock A complete recipe for stochastic gradient mcmc.
\newblock \emph{Advances in Neural Information Processing Systems}, 2015.

\bibitem[Madry et~al.(2018)Madry, Makelov, Schmidt, Tsipras, and
  Vladu]{madry2019deep}
Madry, A., Makelov, A., Schmidt, L., Tsipras, D., and Vladu, A.
\newblock Towards deep learning models resistant to adversarial attacks.
\newblock \emph{Proceedings of the International Conference on Learning
  Representations}, 2018.

\bibitem[Miersemann(2012)]{variation}
Miersemann, E.
\newblock \emph{Calculus of Variations, Lecture Notes}.
\newblock Leipzig University, 2012.

\bibitem[Nguyen et~al.(2018)Nguyen, Li, Bui, and Turner]{nguyen2017variational}
Nguyen, C.~V., Li, Y., Bui, T.~D., and Turner, R.~E.
\newblock Variational continual learning.
\newblock \emph{Proceedings of the International Conference on Learning
  Representations}, 2018.

\bibitem[Pham et~al.(2021)Pham, Liu, Sahoo, and HOI]{pham2021contextual}
Pham, Q., Liu, C., Sahoo, D., and HOI, S.
\newblock Contextual transformation networks for online continual learning.
\newblock \emph{Proceedings of the International Conference on Learning
  Representations}, 2021.

\bibitem[Rahimian \& Mehrotra(2019)Rahimian and
  Mehrotra]{rahimian2019distributionally}
Rahimian, H. and Mehrotra, S.
\newblock Distributionally robust optimization: A review.
\newblock 2019.

\bibitem[Riemer et~al.(2019)Riemer, Cases, Ajemian, Liu, Rish, Tu, and
  Tesauro]{riemer2018learning}
Riemer, M., Cases, I., Ajemian, R., Liu, M., Rish, I., Tu, Y., and Tesauro, G.
\newblock Learning to learn without forgetting by maximizing transfer and
  minimizing interference.
\newblock \emph{International Conference on Learning Representations}, 2019.

\bibitem[Sagawa et~al.(2020)Sagawa, Koh, Hashimoto, and
  Liang]{sagawa2020distributionally}
Sagawa, S., Koh, P.~W., Hashimoto, T.~B., and Liang, P.
\newblock Distributionally robust neural networks for group shifts: On the
  importance of regularization for worst-case generalization.
\newblock \emph{International Conference on Learning Representations}, 2020.

\bibitem[Saha et~al.(2021)Saha, Garg, and Roy]{saha2021gradient}
Saha, G., Garg, I., and Roy, K.
\newblock Gradient projection memory for continual learning.
\newblock \emph{Proceedings of the International Conference on Learning
  Representations}, 2021.

\bibitem[Vinyals et~al.(2016)Vinyals, Blundell, Lillicrap, kavukcuoglu, and
  Wierstra]{NIPS2016_90e13578}
Vinyals, O., Blundell, C., Lillicrap, T., kavukcuoglu, k., and Wierstra, D.
\newblock Matching networks for one shot learning.
\newblock 29, 2016.

\bibitem[von Oswald et~al.(2019)von Oswald, Henning, Sacramento, and
  Grewe]{oswald2019continual}
von Oswald, J., Henning, C., Sacramento, J., and Grewe, B.~F.
\newblock Continual learning with hypernetworks.
\newblock \emph{https://arxiv.org/abs/1906.00695}, 2019.

\bibitem[Wang et~al.(2021)Wang, Duan, Fang, Suo, and Gao]{Wang_2021_ICCV}
Wang, Z., Duan, T., Fang, L., Suo, Q., and Gao, M.
\newblock Meta learning on a sequence of imbalanced domains with difficulty
  awareness.
\newblock In \emph{Proceedings of the IEEE/CVF International Conference on
  Computer Vision}, pp.\  8947--8957, 2021.

\bibitem[Wang et~al.(2022)Wang, Shen, Duan, Zhan, Fang, and
  Gao]{Wang_2022_CVPR}
Wang, Z., Shen, L., Duan, T., Zhan, D., Fang, L., and Gao, M.
\newblock Learning to learn and remember super long multi-domain task sequence.
\newblock In \emph{Proceedings of the IEEE/CVF Conference on Computer Vision
  and Pattern Recognition}, pp.\  7982--7992, 2022.

\bibitem[Welling \& Teh(2011)Welling and Teh]{SGLD11}
Welling, M. and Teh, Y.~W.
\newblock Bayesian learning via stochastic gradient langevin dynamics.
\newblock \emph{Proceedings of the International Conference on Machine
  Learning}, 2011.

\bibitem[Xu et~al.(2020)Xu, Dan, Khim, and Ravikumar]{imbalanceclass}
Xu, Z., Dan, C., Khim, J., and Ravikumar, P.
\newblock Class-weighted classification: Trade-offs and robust approaches.
\newblock \emph{Proceedings of the International Conference on Machine
  Learning}, 2020.

\bibitem[Zenke et~al.(2017)Zenke, Poole, and Ganguli]{zenke2017continual}
Zenke, F., Poole, B., and Ganguli, S.
\newblock Continual learning through synaptic intelligence.
\newblock \emph{https://arxiv.org/abs/1703.04200}, 2017.

\bibitem[Zeno et~al.(2019)Zeno, Golan, Hoffer, and Soudry]{zeno2019task}
Zeno, C., Golan, I., Hoffer, E., and Soudry, D.
\newblock Task agnostic continual learning using online variational bayes.
\newblock \emph{https://arxiv.org/abs/1803.10123}, 2019.

\bibitem[Zhai et~al.(2021)Zhai, Dan, Kolter, and Ravikumar]{zhai2021doro}
Zhai, R., Dan, C., Kolter, J.~Z., and Ravikumar, P.
\newblock Doro: Distributional and outlier robust optimization.
\newblock \emph{Proceedings of the International Conference on Machine
  Learning}, 2021.

\end{thebibliography}
